\documentclass[letterpaper, 10 pt, conference]{ieeeconf}  % Comment this line out if you need a4paper
\usepackage{graphics} % for pdf, bitmapped graphics files
\usepackage{epsfig} % for postscript graphics files
\usepackage{mathptmx} % assumes new font selection scheme installed
\usepackage{amsmath} % assumes amsmath package installed
\usepackage{amssymb}  % assumes amsmath package installed
\usepackage[tight,footnotesize]{subfigure}
\usepackage{leftidx}
\usepackage{cite}
\usepackage{multirow}

\title{\LARGE \bf 3D Maps Registration and Path Planning for Autonomous Robot Navigation}

\author{ \authorblockN{\Large Konstantinos Charalampous$^1$, Ioannis Kostavelis$^1$, Dimitrios Chrysostomou$^2$, \\ Angelos Amanatiadis$^1$ and Antonios Gasteratos$^1$}
\authorblockA{\small $^1$Laboratory of Robotics and Automation, Department of Production and Management Engineering,\\
Democritus University of Thrace,
Vas. Sofias 12, GR-67100
Xanthi, Greece\\
\{kchara, gkostave, dchrisos, agaster\}@pme.duth.gr. 
\{aamanat\}@ee.duth.gr.\\
}\\
\authorblockA{\small $^2$Robotics and Automation Group,
 Department of Mechanical \& Manufacturing Engineering,\\
Aalborg University,
Fibigestraede 16, DK-9220
Aalborg East, Denmark\\
dimi@m-tech.aau.dk \\
}\\
}

\begin{document}
\maketitle
\thispagestyle{empty}
\pagestyle{empty}

%%%%%%%%%%%%%%%%%%%%%%%%%%%%%%%%%%%%%%%%%%%%%%%%%%%%%%%%%%%%%%%%%%%%%%%%%%%%%%%%
\begin{abstract}
Mobile robots dedicated in security tasks should be capable of clearly perceiving their environment to competently navigate within cluttered areas, so as to accomplish their assigned mission. The paper in hand describes such an autonomous agent designed to deploy competently in hazardous environments equipped with a laser scanner sensor. During the robot's motion, consecutive scans are obtained to produce dense 3D maps of the area. A 3D point cloud registration technique is exploited to merge the successively created maps during the robot's motion followed by an ICP refinement step. The reconstructed 3D area is then top-down projected with great resolution, to be fed in a path planning algorithm suitable to trace obstacle-free trajectories in the explored area. The main characteristic of the path planner is that the robot's embodiment is considered for producing detailed and safe trajectories of $1$ $cm$ resolution. The proposed method has been evaluated with our mobile robot in several outdoor scenarios revealing remarkable performance.
\end{abstract}
%%%%%%%%%%%%%%%%%%%%%%%%%%%%%%%%%%%%%%%%%%%%%%%%%%%%%%%%%%%%%%%%%%%%%%%%%%%%%%%%
\section{Introduction}
The development of efficient methods that allow reliable robot navigation and deployment in hazardous environments, remains an active research topic\cite{liu2013robotic}. The proof is that the robotic urban search and rescue community bears a handful of robots in the rescue and recovery operations of many recent devastations, such as the 2004 Mid Niigata earthquake in Japan\cite{meguro2006disaster}. Moreover, in\cite{gianni2011}, the framework of the NIFTi project is presented, which took place after a sequence of earthquakes in the region of Emilia-Romagna in Northern Italy. This framework presents the successful deployment of a team of humans and robots that produced significant achievements in the domain of urban search and rescue.% Additionally, in the work described in \cite{baudoin2009view}, robots equipped with sensors can detect the presence of chemicals, assisting in such way the fire-fighting services and crisis management units. Other areas of applications where robotic technologies can be utilized are evacuation assistance tasks\cite{shell2005} or bomb disposals~\cite{beltran2007methods}.

%\begin{figure}
%	\begin{center}{\includegraphics[width=.3\textwidth]{images/robot.pdf}
%		\label{rob1}}
%	\end{center}
%\caption{The robotic platform ERA-Mobi equipped with a SICK LMS 500 PRO laser sensor and a SCHUNK PW-70 pan-tilt unit.}
%\label{robot}
%\end{figure}

The accurate mobile robot navigation is of great importance especially when it concerns applications in the rescue domain. In \cite{gianni2011}, an integrated system of high level planning and execution with incoming perceptual information from vision, SLAM and topological map segmentation is proposed. This paper describes a system that focuses on two different objectives \cite{aamanat}, which are essential for the navigation of mobile robots in unexplored hazardous environments:
%\begin{itemize}
(i) the development of an accurate \emph{3D reconstruction} and \emph{registration} algorithm suitable to produce dense 3D maps and precise estimations of the robot's motion % It employs the Fast Point Feature Histograms (FPFH) ~\cite{rusu2009fast} for the initial alignment of the 3D maps. However, since this method has been proved to exhibit poor performance in open surfaces, an additional Iterative Closest Point (ICP) ~\cite{yang1992object} routine is applied after the rough alignment, so as to refine the resulted 3D map.}
and (ii) the integration of a \emph{path planning algorithm} within the resulted 3D map in order to produce a collision free trajectory. It utilizes the D* Lite \cite{koenig2005fast} path planning algorithm including specific optimizations to take into consideration the robot's embodiment during the trajectory calculation.
% The aforementioned modules are integrated seamlessly in a robotic platform, depicted in Fig.~\ref{robot}.
%    \end{itemize}
%The efficiency of the proposed system has been thoroughly examined in outdoor scenarios and exhibited remarkable performance during our robot's deployment in unknown environments.
%The rest of the paper is organized as follows; In Sec.~\ref{related} the related work is outlined, while in Sec.~\ref{3D:perc} the algorithm for the perception of the 3D scene is presented. Sec.~\ref{robot:navigation} describes the robot navigation scheme. The experimental results are summarized in Sec.~\ref{exp}, while conclusions are drawn in Sec.~\ref{concl}.

Several approaches exist for deriving a collision free trajectory utilizing point clouds especially for outdoors scenarios.
%However, most of the methodologies are not suitable for real time operation due to the high computational burden. The latter is an after-effect of the density of the point clouds, as well as the size of the derived maps.
%For indoor environments, Biswas et al. \cite{biswas2012depth} suggested the Fast Sampling Plane Filtering (FSPF) algorithm, which samples the derived point cloud of a depth image to produce a set of points that correspond to planes accompanied by the respective parameters, namely the normals and the offsets. The proposed filtering considerably decreases the volume of the data, while at the same time rejects non-planar objects. The FSPF method compiles a list of 3D points, plane normals and outliers, where outlier points are considered to be the non-planar points. This filtering procedure encapsulates the RANSAC algorithm, which is locally applied in a pre-defined search window. Moreover, the localization algorithm is further enhanced by an observation model responsible for matching the extracted plane points in 2D maps, thus allowing the reuse of the previously acquired maps.\par
 The work in \cite{wurm2009improving} proposes a framework for street robot navigation by detecting low vegetation based on 3D scans. This method classifies three-dimensional scans of the robot environment in to two classes, namely the flat vegetation and the streets. A support vector machine (SVM) classifier is utilized on the laser scanner remission values which rely on the material of the measured surface, the distance and the angle of incidence.
% The trained classifier is then utilized for enhancing the navigation adequacy of the system.
 The work in \cite{stoyanov2010path} proposed the usage of 3D Normal Distributions Transform (3D-NDT) as a path planning data structure. Although NDT was initially presented as a 2D laser scan registration technique, the work in \cite{stoyanov2010path} proposed modifications to derive an accurate path planner, namely a collision check routine and a check whether it is feasible for the robot to move from one cell having a certain configuration to another.%The last modification affects the cost function between two corresponding cells which becomes proportional to their Euclidean distance, in order to deal with the non-uniform grid derived by the 3D-NDT transform. Zhuang et al. \cite{5553478} proposed an edge-feature based ICP algorithm for 3D scene registration and reconstruction. The motion planning routine which is based on metric and elevation information derives the corresponding road map, while the optimal trajectory is computed by considering multiple constrains.

\section{3D Scene Perception}\label{3D:perc}

There are several robot set ups and registration algorithms cited in the literature for generating an accurate 3D scene representation. Speed and accuracy are the two most critical aspects in such configurations. The set up followed in this work as well as the registration technique are described in detail in this section.
%\subsection{Robot Set Up}\label{robot:setup}
The platform is the wheeled robot ERA-Mobi.
% The dimensions of the platform are 40 cm in length, 41 cm in width and 15 cm in height.
%Our differential wheeled robot has a maximum speed of $2$ ${m}/{sec}$ and the respective payload is $20$ $Kg$.  The on-board pc has a 2.0 GHz Intel Core2Duo processor and 4GB of RAM. Additionally, a SCHUNK PW-70 servo-electric rotary pan-tilt actuator is utilized in the proposed work. The rotation angle for the pan axis is $360^{\circ}$, while the maximum acceleration and angular velocity are $1440^{\circ}/{sec^{2}}$ and $360^{\circ}/{sec}$, respectively.
 Regarding the point cloud formation, the SICK LMS500 PRO 2D laser scanner was mounted on the PTU and configured for vertical scanning.
%In more detail, the field of view is $190^{\circ}$, the angular resolution varies within the interval $[0.1667^{\circ}-1^{\circ} ]$, the scanning frequency alters in the $[25-100]$ $Hz$ interval, while the scanning range is up to $80$ $m$. 
%According to manufacturer's specifications the standard deviation of the measurements is of $0.01$ $m$. SICK's angular resolution was set to $0.1667^{\circ}$ and PTU's angular velocity of the pan axis at $3^{\circ}/sec$. The 3D scan is produced by performing a $360^{\circ}$ scan sweep.
 Last, a 3DoF orientation tracker from Xsens accompanies the system enabling the orientation measurement of the robotic platform.

% The robot operating system (ROS) was adopted as the most suitable operating system since it offers driver support of the aforementioned devices, including the robot platform. While the different devices do not communicate with each other and also there is not an external triggering mechanism to synchronize them, ROS undertakes this task using timestamps to the messages published from each device. besl1992method

%\subsection{Point Cloud Registration}\label{point:cloud}
%
The point cloud registration is accomplished into two distinct steps. First, a rough estimation of the transformation matrix is given using the FPFH \cite{rusu2009fast} features followed by the ICP algorithm as a refinement step.
FPFH are multi-dimensional features describing the geometry of a point belonging to a 3D point cloud allowing the on-line calculation of those features, making it suitable for on-line applications.
The required rigid body transformation typically should conform with a sum of quadratic differences minimization criterion, resulting to a singular value decomposition (SVD) optimization problem. By applying the motion transformation on the respective 3D point clouds we obtain a rough alignment and, as a result, the 3D map retains erroneous registrations. It is worth mentioning that for this rough alignment an initialization step takes place regarding the orientation of the robot by exploiting the orientation tracker device. Hence, the initially transformed point clouds are considered for the correction of the motion estimation. The most commonly used algorithm to fine register the 3D point clouds is the ICP one.
However, the novelty of the proposed work is that our ICP algorithm considers only the points that belong to specific geometric surfaces in consecutive time instances. The successive point clouds share great amount of spatial proximity, due to the fact that a coarse alignment occurred during the motion estimation procedure. The benefit from this procedure is twofold: firstly we avoid multiple iterations restricting the rigid body transformation search by one order of magnitude in calculation time and, secondly, we increase the likelihood to achieve an accurate solution. These advantages are feasible due to the fact that the considered points are contained in two successively observed scenes. Concerning the two successive 3D point clouds $\leftidx{^t}{\mathbf{P'}}$ and $\leftidx{^{t+1}}{\mathbf{P'}}$, we utilize a point-to-plane ICP algorithm\cite{yang1992object}, which seeks for a transformation $K$, that registers the two point clouds.

%The output of the point-to-plane ICP algorithm is a transformation $K = [\leftidx{_{t+1}^t}{R_{ICP}}, \leftidx{_{t+1}^t}{T_{ICP}}]$ that aligns the two successive point clouds. The transformation $K$ is combined with the initial estimation, as resulted from the FPFH registration and a refined estimation of the robot's pose is obtained, i.e. $\leftidx{_{t+1}^t}{R_{ref}} = \leftidx{_{t+1}^t}{R} \cdot \leftidx{_{t+1}^t}{R_{ICP}}$ and $\leftidx{_{t+1}^t}{T_{ref}} = \leftidx{_{t+1}^t}{T}+\leftidx{_{t+1}^t}{T_{ICP}}$. The calculated $\leftidx{_{t+1}^t}{R_{ref}}$ and $\leftidx{_{t+1}^t}{T_{ref}}$ are then introduced in the accurate registration of the successive 3D point clouds. This procedure is performed separately in each time step and, hence, the 3D map of the explored area is constructed incrementally. %Comparing Fig. \ref{fig:fpfh_sole} where the point clouds have been merged by applying only the initial registration estimation with the output of the ICP refinement step, as depicted in Fig. \ref{fig:fpfh_icp}, it can be shown that the second framework produces more accurate and consistent results. That is, the aforementioned refinement invokes essentially on the 3D point cloud registration.
%
\section{Robot Navigation}\label{robot:navigation}
%This section describes the derivation methodology of the 2D map from the registered 3D point cloud, the robot embodiment feature, as well as the path planning procedure.
%
%\begin{figure*}[!ht]
%\hfil
%\begin{center}
%{
%	\subfigure[]{\includegraphics[width=2.25in]{images/indoors_1.pdf} \label{fig:indoors_1}}
%	\hfil
%	\subfigure[]{\includegraphics[width=2.25in]{images/indoors_2.pdf} \label{fig:indoors_2}}
%     \hfil
%	\subfigure[]{\includegraphics[width=2.25in]{images/indoors_3.pdf} \label{fig:indoors_3}}
%}
%\caption{a) Registered point clouds from indoors exploration (the ceiling has been removed for visualization purposes), b) the derived map after the top-down projection, c) the resulting trajectories with and without the consideration of the embodiment.}
%\label{fig:Indoors_case}
%\end{center}
%\end{figure*}
%
%
\begin{figure*}[!ht]
\hfil
\begin{center}
{
	\subfigure[]{\includegraphics[width=2.25in]{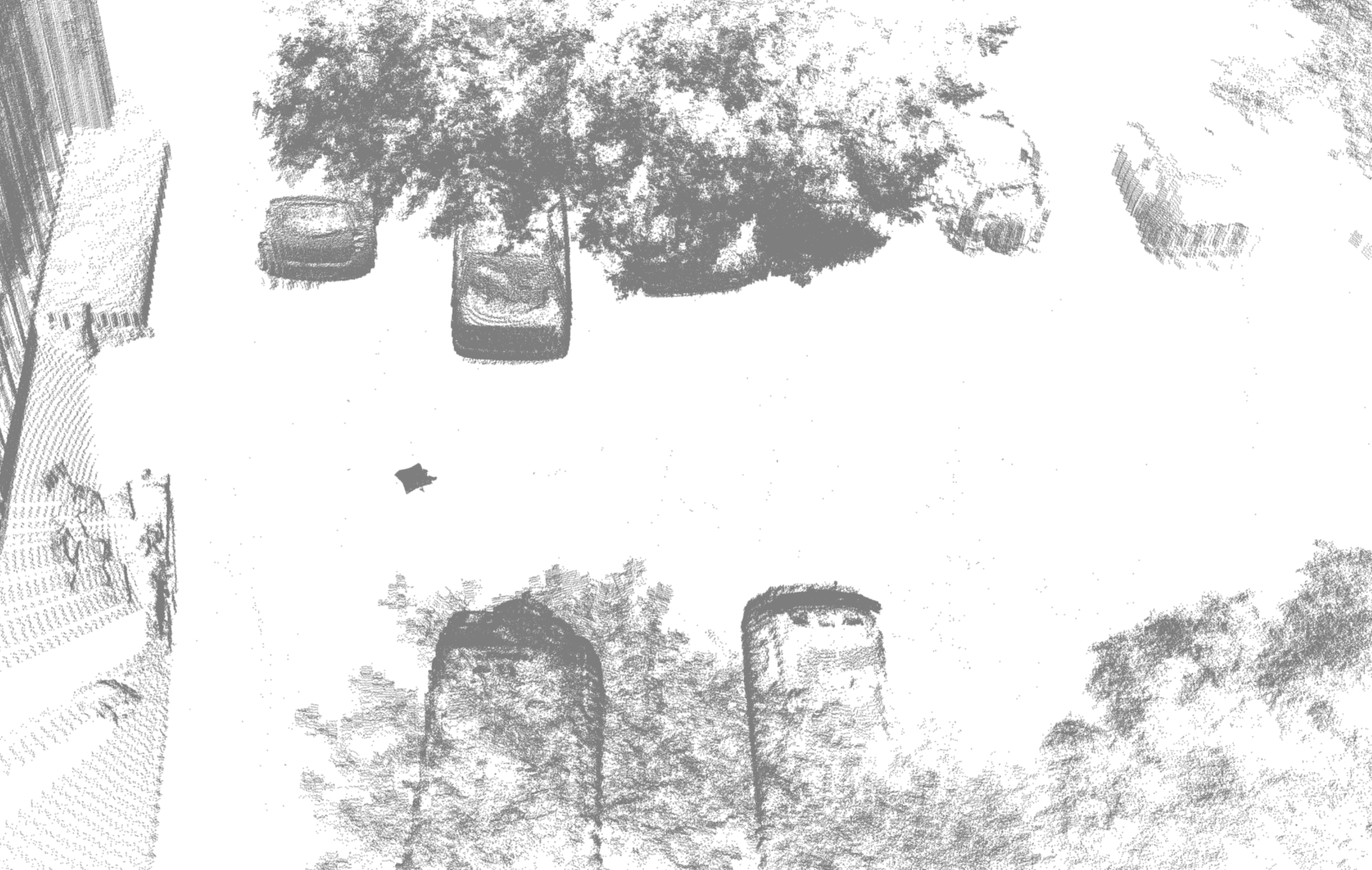} \label{fig:outdoors_1}}
	\hfil
	\subfigure[]{\includegraphics[width=2.25in]{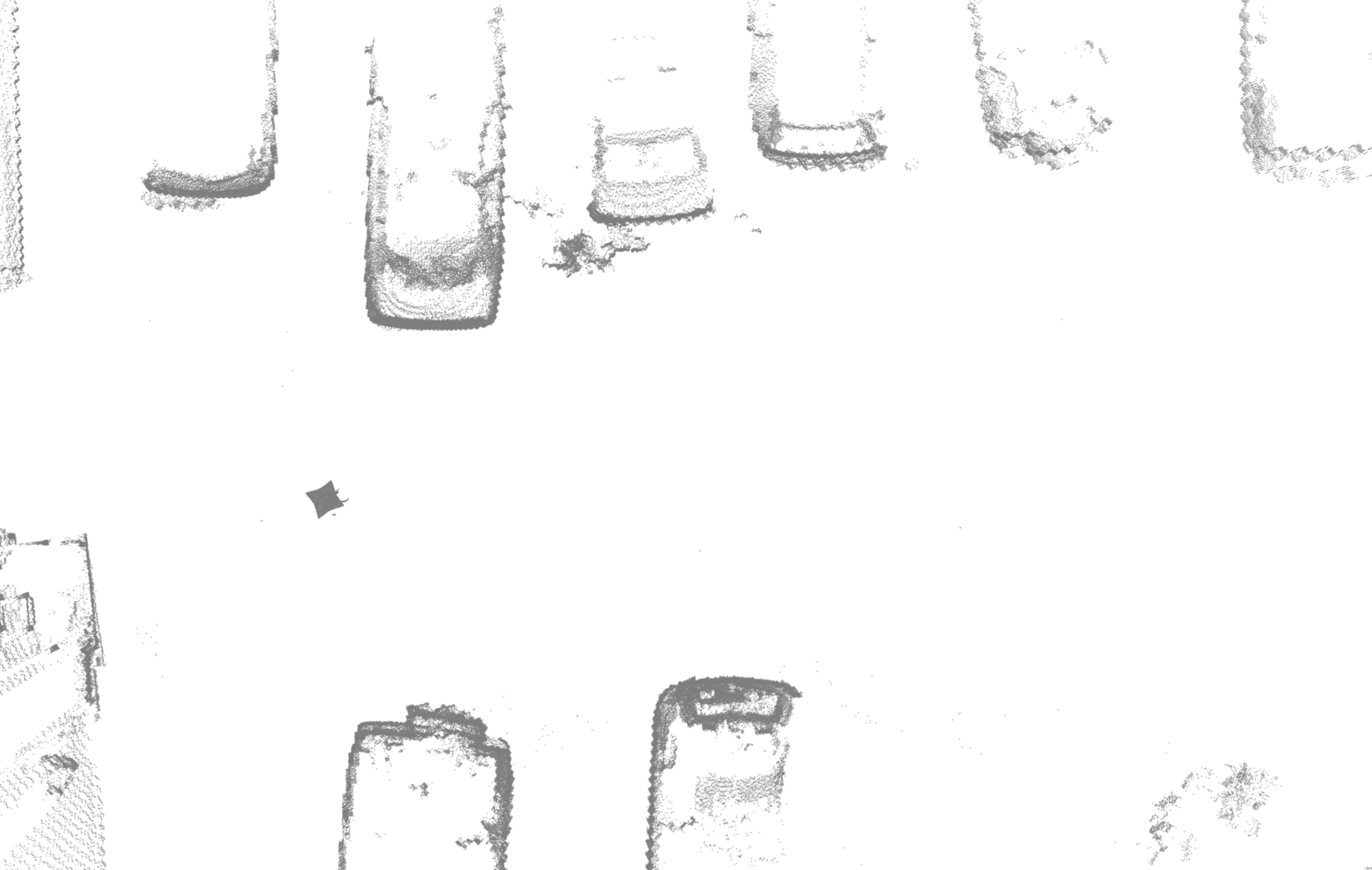} \label{fig:outdoors_2}}
     \hfil
	\subfigure[]{\includegraphics[width=2.25in]{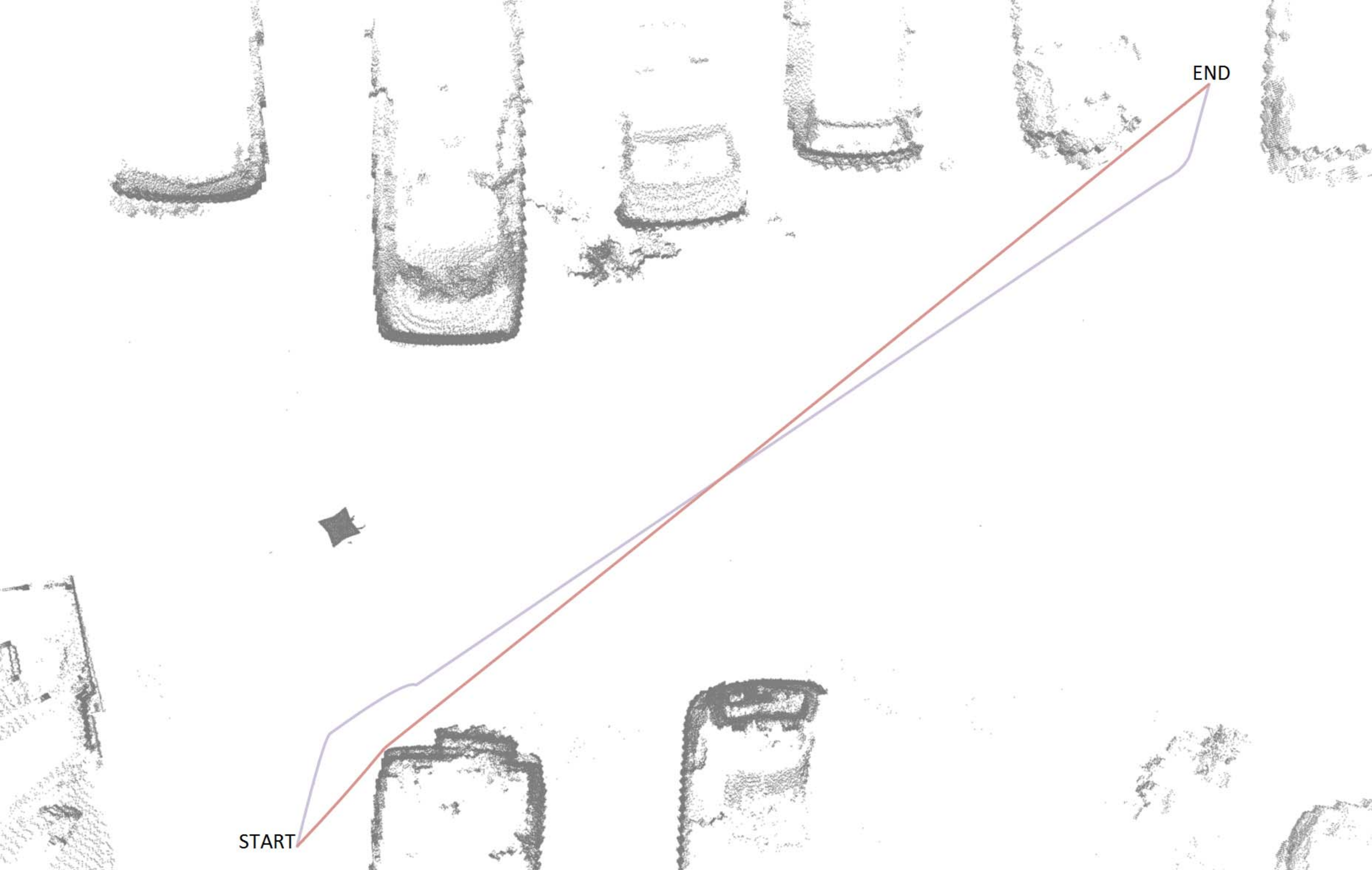} \label{fig:outdoors_3}}
}
\caption{a) Registered point clouds from outdoors exploration, b) the derived map after the top-down projection, c) the resulting trajectories with and without the consideration of the embodiment.}
\label{fig:Outdoors_case}
\end{center}
\end{figure*}
\begin{figure*}[!ht]
\hfil
\begin{center}
{
	\subfigure[]{\includegraphics[width=2.25in]{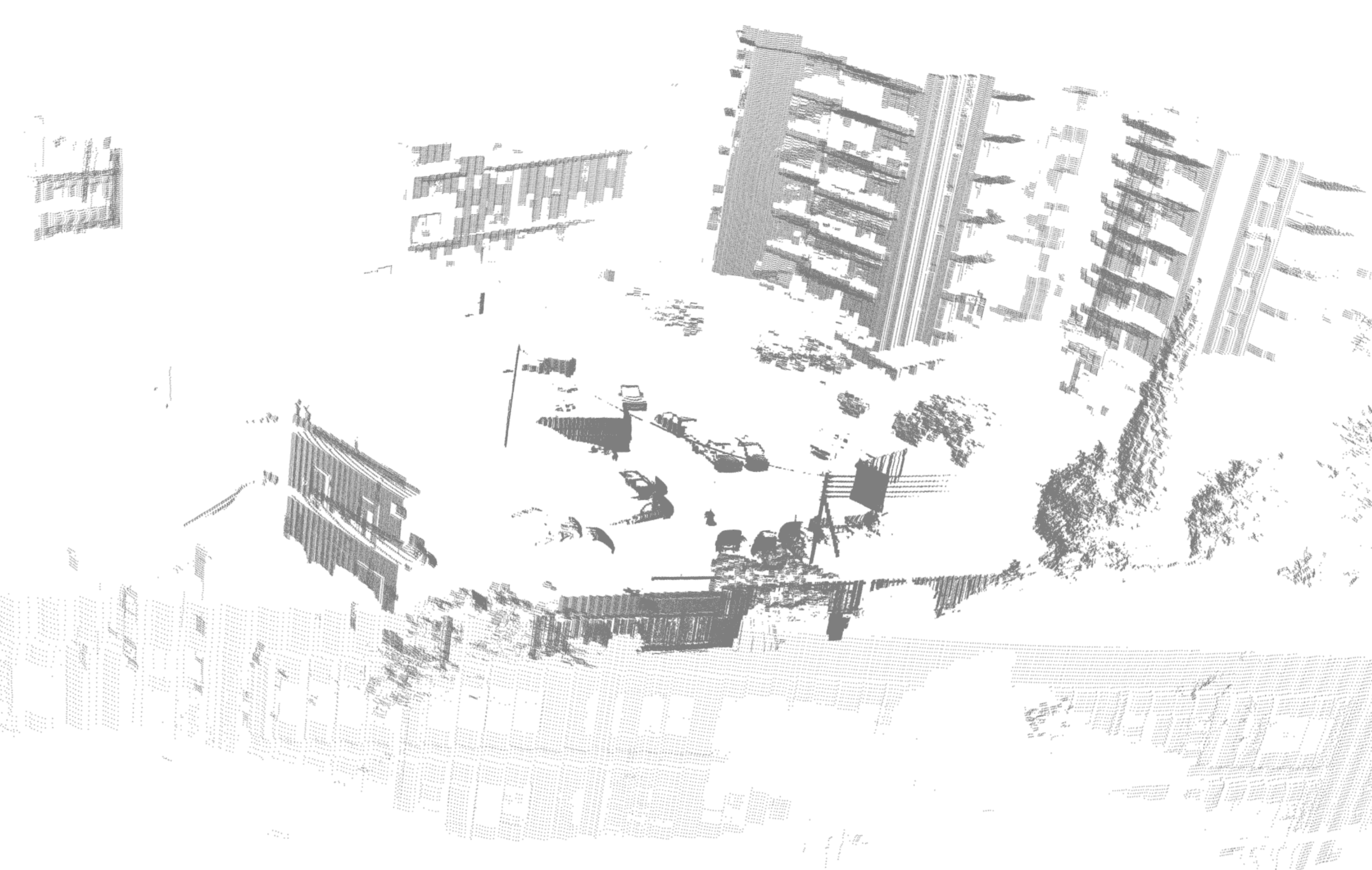} \label{fig:outdoors2_1}}
	\hfil
	\subfigure[]{\includegraphics[width=2.25in]{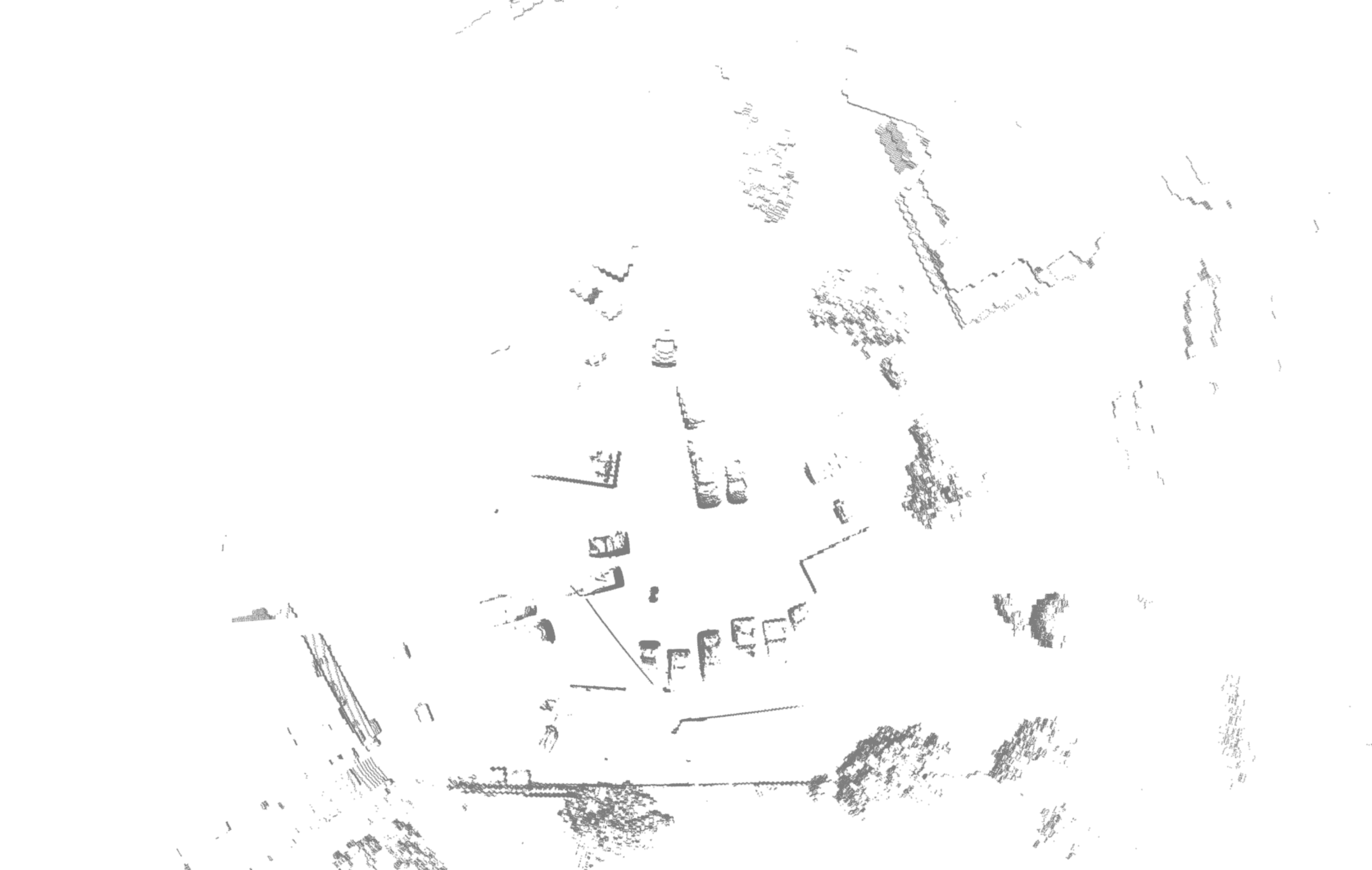} \label{fig:outdoors2_2}}
     \hfil
	\subfigure[]{\includegraphics[width=2.25in]{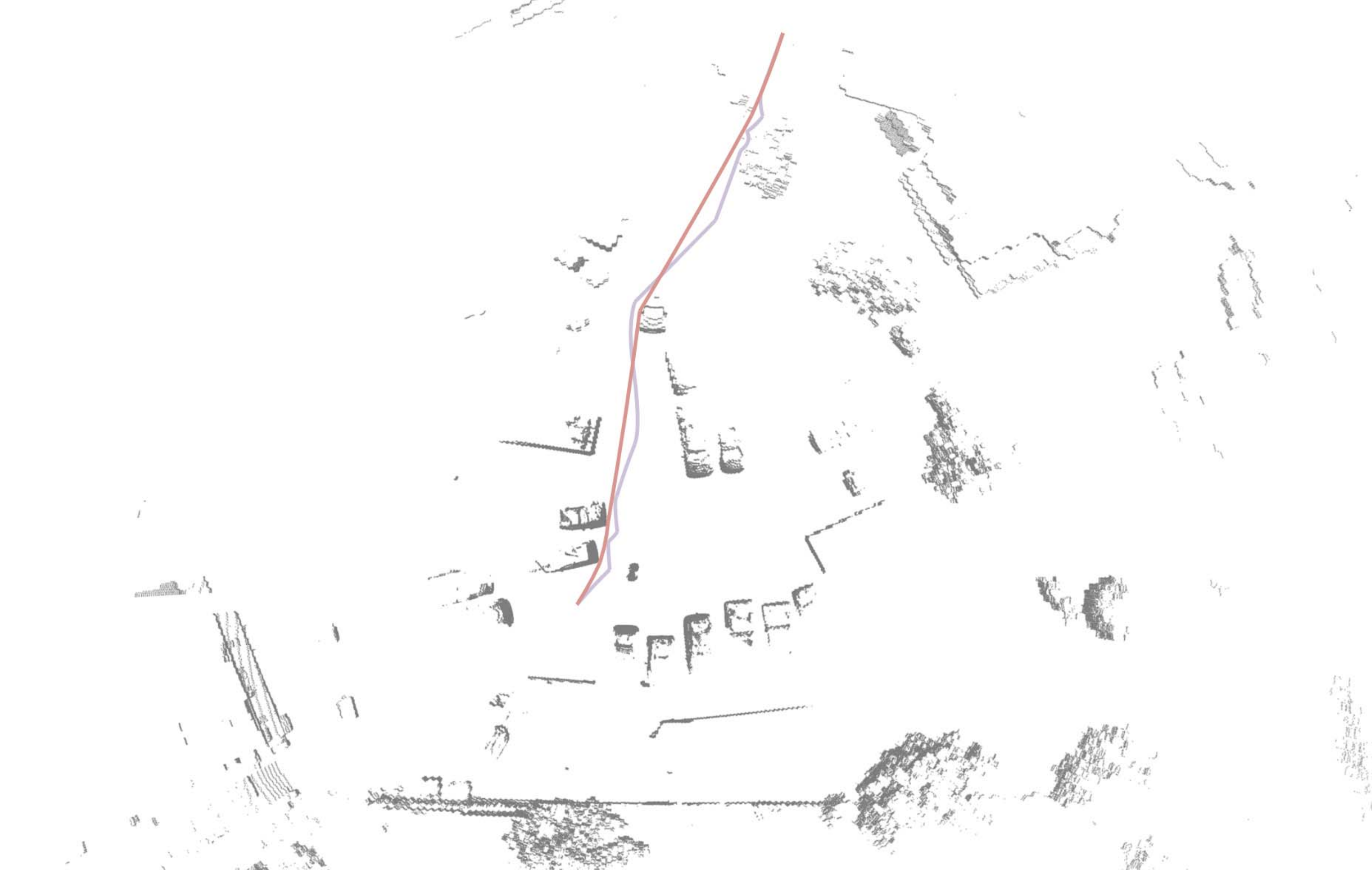} \label{fig:outdoors2_3}}
}
\caption{a) Registered point clouds from outdoors exploration in a larger field, b) the derived map after the top-down projection, c) the resulting trajectories with and without the consideration of the embodiment.}
\label{fig:Outdoors_case2}
\end{center}
\end{figure*}
%
% \subsection{Top Down Projection and Filtering}\label{proj:filter}
Once the registration procedure between the 3D point clouds has been completed, the derived transformation is applied and a new point cloud results by merging the previous ones. The new cloud contains various points that have identical coordinates and, thus, a filtering procedure is performed. The voxelized grid approach followed leads to a significant reduction to the number of points, facilitating the manipulation of the data. The voxels have an edge of $1$ $cm$, thus the points within each voxel are approximated by their centroid. The point cloud is referred to the SICK laser scanner, thereafter the floor is removed by applying the RANSAC plane detection algorithm. It is expected that the largest plane in the examined scene would be the floor and, therefore, the RANSAC algorithm operates vertical to the robot axis. This interval is deduced by the following two considerations: $i)$ these points should have zero values on the corresponding axis, and while the robot platform can shift over $2$ $cm$ obstacles, points within the range of $[0,2]$ $cm$ should be also removed $ii)$ taking into account the standard deviation ($1$ $cm$) as indicated by the manufacturer, the deviation is appended on both sides of the aforementioned interval. Likewise, points having respective height value over $1.5$ $m$ are also removed since it is impossible for the robot to collide with them. The remaining points are top-down projected and a 2D map of a centimeter accuracy is acquired.

All points of the 2D map are declared as non-traversable in the path planning algorithm. At this point, if a path planning procedure was executed, then, in order to derive the shortest path,	 the successive points of this path would probably have to pass very close or next to the obstacles. Regarding the path planning procedure itself, this would be a correct path, since the robot is assumed to be punctual. By considering the robot's embodiment, avoidance of such paths is attained by assigning certain cost values to the neighbor points of the non-traversable ones. In more detail, the ERA-Mobi platform's length and width are $40$ $cm$ and $41$ $cm$, respectively. That is, the most distant point from the center of the platform is at $28.64$ $cm$ and, since the resolution of the map is of the order of one centimeter we round that value to $29$ $cm$. For every non-traversable point $ob^{i}$, its neighbor cells that abstain up to $29$ $cm$ acquire penalty cost values according to the 2D Gaussian function having standard deviation $\sigma_{x}, \sigma_{y} = 1$, i.e. $f(x,y)=e^{-((x-ob^{i}_{x})^{2}/2+(y-ob^{i}_{y})^{2}/2)}$, where $ob^{i}_{x},ob^{i}_{y}$ are the coordinates of $ob^{i}$. If two or more non-traversable points have overlapping neighbors, then these cost values of those points are accumulated. This process creates a trade-off between the shortest and the safest path. It is obvious that the latter procedure can be adopted to any platform, resulting a different size of neighborhood.

 The derivation of such a detailed map in conjunction with the cost assigning process enables the extraction of a particularly detailed and safe path. Such a dense path is a notably approximation of a continuous trajectory. Even if the D* Lite operates in a discrete space, the resolution of the implementation produces sufficient accuracy to navigate within the environment. The enhanced map is provided as an input to the D* Lite method. The latter treats the problem as a graph-traversal one and due to the resolution of the map the distance between two nodes in the graph corresponds to $1$ $cm$ in the real world.
%
%\subsection{Path Planning}\label{path:planning}
%%
%The enhanced map described in Sec. ~\ref{proj:filter} is provided as an input to the D* Lite method. The latter treats the problem as a graph-traversal one and due to the resolution of the map the distance between two nodes in the graph corresponds to $1$ $cm$ in the real world.
%D* Lite is a fast path planning and re-planning algorithm suitable for goal-directed robot navigation in partial known or unknown terrain. D* Lite constitutes an extension of Lifelong Planning A* (LPA*)~\cite{koenig2001incremental}, \cite{charalampous2012efficient} and one of its most significant additions is the variation where the target position changes dynamically during the re-planning phases. Due to the fact that D* Lite expands LPA*, it also acquires the entire set of attributes that LPA* entails and its expansion capabilities as well. Compared to other methodologies, such as the well known D*~\cite{stentz1995focussed}, D* Lite is simpler, can be rigorously analyzed, can be extended in various aspects and its efficiency is at least equal to the one of D*. Regarding the simplicity, D* Lite utilizes a single tie-breaking rule for the comparison of the priorities, making their maintenance an easy task.
%
\section{Experimental Results}\label{exp}

Considering the first outdoor scenario, the registration procedure is illustrated in Fig. \ref{fig:outdoors_1}, where the transformation matrix between the point clouds is correctly computed while Fig. \ref{fig:outdoors_2} depicts the resulted 2D map. The average time of the rough registration procedure using the FPFH features was $38.4$ $sec$, while the refinement using the ICP lasted $2.9$ $sec$. The computation of the path has mean time $2.4$ $sec$, while Fig. \ref{fig:outdoors_3} illustrates the paths derived by D* Lite, where the purple line indicates the path in which the embodiment of the robot was considered, and the red one shows the route without taking into account the respective embodiment. The second outdoor scenario took place in a significantly larger area. The distance between the two $360^{0}$ laser scans was $35$ $m$, yet the registration of the point clouds was legitimate as illustrated in Fig. \ref{fig:outdoors2_1}. It is worth mentioning that the computational time for the registration remained almost the same as in the previous outdoor scenario. The respective 2D map is depicted in Fig. \ref{fig:outdoors2_2} where the ground was also accurately removed. The calculation of the path lasted almost the same as in the later case, and again, as illustrated in Fig. \ref{fig:outdoors2_3}, the robot does not navigate nearby obstacles, offering a safer path to the goal.
 Table \ref{tab:time} summarizes the execution times of each routine separately for the aforementioned scenarios.
\begin{table}[!ht]
\caption{Average time of execution in five runs.} %title of the table
\centering % centering table
\begin{tabular}{| p{2cm}| p{2.5cm}| p{2.5cm}|} % creating eight columns
\hline %inserting double-line
                & First case $\pm$ std & Sec. case $\pm$ std\\ [0.5ex]
\hline % inserts single-line
FPFH      & 38.4  $\pm$ 6.4       & 38.7  $\pm$ 6.1\\
ICP          & 2.9     $\pm$ 1.7    & 3.0     $\pm$ 1.8 \\
D* Lite   & 2.4     $\pm$ 0.3    & 2.5     $\pm$ 0.4 \\
\hline % inserts single-line
\end{tabular}
\label{tab:time}
\end{table}
\section{Conclusion}\label{concl}
In this work, a novel integrated system for autonomous robot navigation has been presented utilizing only a laser scanner sensor for the 3D perception of the environment. For every robot motion update, the system generates a 3D map of the environment and by utilizing accurate registration techniques, a consistent 3D map of the explored area is produced. The extracted 3D map is further exploited by the robot for estimating detailed paths within the explored scene for safe navigation. 
%The path planner operates in continuous space by considering also the robot's embodiment generating safe navigation trajectories. The aforementioned modules have been tested both as stand-alone routines and as an integrated framework which can operate seamlessly in an autonomous fashion.
 The proposed system is a complete navigation framework that can be deployed in hazardous or post-disaster situations aiming to assist rescue activities.

 \section*{Acknowledgment}
% optional entry into table of contents (if used)
%\addcontentsline{toc}{section}{Acknowledgment}
This work was fully supported by the E.C. under the FP7 research project for \textit{The Autonomous Vehicle Emergency Recovery Tool to provide a robot path planning and navigation tool}, "AVERT", FP7-SEC-2011-1-285092.

\bibliographystyle{IEEEtran}
\bibliography{IEEEabrv,refs}

\end{document}